\documentclass[lettersize,journal]{IEEEtran}
\usepackage{amsmath,amsfonts}
\usepackage{algorithmic}
\usepackage{algorithm}
\usepackage{array}
\usepackage[caption=false,font=normalsize,labelfont=sf,textfont=sf]{subfig}
\usepackage{textcomp}
\usepackage{stfloats}
\usepackage{url}
\usepackage{verbatim}
\usepackage{graphicx}
\usepackage{cite}
\usepackage{amsmath}
\usepackage{placeins}
\usepackage{hyperref} 
\usepackage{academicons} 
\usepackage{svg}

\begin{document}

\title{Deep-BrownConrady: Prediction of Camera Calibration and Distortion Parameters Using Deep Learning and Synthetic Data}

\author{
    Faiz Muhammad Chaudhry\href{https://orcid.org/0009-0008-8921-924X}{\includegraphics[width=1em]{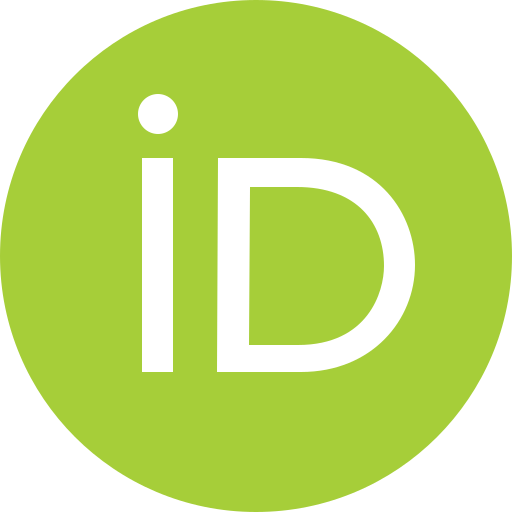}}, 
    Jarno Ralli\href{https://orcid.org/0000-0001-7110-4294}{\includegraphics[width=1em]{img/orcid.png}}, 
    Jerome Leudet\href{https://orcid.org/0009-0006-3135-9034}{\includegraphics[width=1em]{img/orcid.png}}, 
    Fahad Sohrab\href{https://orcid.org/0000-0001-6526-3811}{\includegraphics[width=1em]{img/orcid.png}}, Member, IEEE, 
    Farhad Pakdaman\href{https://orcid.org/0000-0001-6526-3811}{\includegraphics[width=1em]{img/orcid.png}}, Member, IEEE, 
    Pierre Corbani\href{https://orcid.org/0009-0006-1478-9809}{\includegraphics[width=1em]{img/orcid.png}}, and
    Moncef Gabbouj\href{https://orcid.org/0000-0002-9788-2323}{\includegraphics[width=1em]{img/orcid.png}}, Fellow, IEEE

\thanks{Manuscript created December, 2024; This work was funded by AILiveSim Ltd. Faiz Muhammad Chaudhry, Jarno Ralli, Jerome Leudet and Pierre Corbani are with AILiveSim Ltd., 00180 Helsinki, Finland (email: faiz@ailivesim.com; jarno@ailivesim.com; jerome@ailivesim.com; pierre@ailivesim.com).\\
Fahad Sohrab, Farhad Pakdaman and Moncef Gabbouj are with the Faculty of Information Technology and Communication Sciences, Tampere University, 33720 Tampere, Finland (e-mail: fahad.sohrab@tuni.fi; farhad.pakdaman@tuni.fi; moncef.gabbouj@tuni.fi).}}

\markboth{IEEE TRANSACTIONS ON AUTOMATION SCIENCE AND ENGINEERING}%
{Shell \MakeLowercase{\textit{et al.}}: A Sample Article Using IEEEtran.cls for IEEE Journals}

\maketitle
\begin{abstract}
This research addresses the challenge of camera calibration and distortion parameter prediction from a single image using deep learning models. The main contributions of this work are: (1) demonstrating that a deep learning model, trained on a mix of real and synthetic images, can accurately predict camera and lens parameters from a single image, and (2) developing a comprehensive synthetic dataset using the AILiveSim simulation platform. This dataset includes variations in focal length and lens distortion parameters, providing a robust foundation for model training and testing. The training process predominantly relied on these synthetic images, complemented by a small subset of real images, to explore how well models trained on synthetic data can perform calibration tasks on real-world images. Traditional calibration methods require multiple images of a calibration object from various orientations, which is often not feasible due to the lack of such images in publicly available datasets. A deep learning network based on the ResNet architecture was trained on this synthetic dataset to predict camera calibration parameters following the Brown-Conrady lens model. The ResNet architecture, adapted for regression tasks, is capable of predicting continuous values essential for accurate camera calibration in applications such as autonomous driving, robotics, and augmented reality.
\end{abstract}


\section*{Note to Practitioners}
This paper introduces a deep learning approach to predict camera calibration and distortion parameters directly from a single image, addressing the limitations of traditional methods that require structured calibration objects and multiple images. Using synthetic datasets generated with a simulation platform, the model predicts essential parameters such as field of view, principal points, and Brown-Conrady distortion coefficients. A key innovation is incorporating image size into the learning process, enabling the model to generalize well to real-world scenarios. This approach simplifies the calibration process, making it suitable for dynamic and unstructured environments, such as autonomous driving and robotics, where traditional calibration methods are not feasible. As an important result, the proposed method enables the use of synthetic data to overcome data scarcity in real-world applications, by adapting to the camera parameters of the underlying physical system. By leveraging synthetic data and deep learning, the method offers a modern, flexible alternative that practitioners can adopt to enhance camera calibration workflows in real-world applications. Extensive experiments are reported which showcase a successful solution based on the famous Brown-Conrady camera lens model, and are validated on a real-world dataset. We believe similar methodology can be used and extended as future work, to enable camera parameter estimation, and the use of synthetic data, for other camera models.
\section{Introduction}
\IEEEPARstart{C}{amera} calibration and distortion correction are fundamental processes in computer vision applications, ensuring geometrically accurate visual data for reliable analysis and decision-making. Autonomous vehicles, Augmented Reality (AR) systems, and robotic vision heavily depend on these processes for optimal performance \cite{yan2023sensorx2car, Whitaker, BALANJI2022102248}. Accurate calibration ensures that the visual data from the camera matches real-world geometry, while distortion correction eliminates artifacts that could otherwise cause errors in tasks like object detection, navigation, and virtual object overlays.

The importance of precise calibration extends beyond conventional vision systems. For instance, robotic surgery systems such as the Da Vinci platform rely on robust camera-robot calibration to ensure accurate hand-eye coordination during minimally invasive procedures \cite{ozguner2020davinci}. This highlights the critical role of calibration in diverse applications, from healthcare robotics to autonomous navigation.

A significant challenge in camera calibration is the absence of specific calibration images in most publicly available datasets, which complicates the accurate estimation of camera and lens distortion model parameters. Traditional methods often rely on multiple images of a calibration object, such as a chessboard pattern, to estimate these parameters \cite{Gasparini2009, MeiRives2007, Scaramuzza2006}. However, such data is not available in most publicly available datasets such as COCO, ImageNet and Open Images \cite{cocodataset,deng2009imagenet, openimages}.

The main objective of this study is to predict camera calibration and lens distortion parameters from a single image, enabling more effective use of computer vision in real-world applications and making it possible to utilize datasets where calibration data is unavailable. We achieve this by generating a comprehensive synthetic dataset using the AILiveSim simulation platform \cite{jerome}, which includes diverse camera settings such as varying fields of view (FOV) and lens distortions. Without loss of generality, we chose the Brown-Conrady \cite{Brown1971} model to represent lens distortions. This model is well-suited for linear cameras when distortions are not expected to be severe. The primary camera parameters targeted for prediction in this study include the horizontal field of view (H-FOV), principal points (\(c_x, c_y\)), and the lens distortion coefficients (\(k_1, k_2, k_3, p_1, p_2\)). These parameters directly influence how the scene observed by the camera is projected onto the image plane, making them vital for applications such as autonomous driving, robotics, and augmented reality.

The proposed approach employs a ResNet-based deep learning model, adapted for regression tasks to estimate these intrinsic camera parameters from the synthetic images. The model is improved by incorporating image size and aspect ratio into the learning process, significantly enhancing its ability to generalize across different sensor sizes and types. This innovation addresses the challenge of varying image resolutions that can affect model accuracy.

We validate the effectiveness of our approach through extensive experiments, demonstrating the model’s ability to generalize from synthetic data to real-world scenarios using the KITTI dataset \cite{kitti}. The results indicate that our method performs robustly in predicting camera calibration parameters, bridging the gap between synthetic training environments and practical deployment.

This study makes several key contributions to the field of camera calibration and distortion correction:

\begin{enumerate}
    \item \textit{Synthetic Dataset Creation:} A comprehensive synthetic dataset was generated using the AILiveSim platform \cite{jerome}, incorporating various FOV and principle point settings and Brown-Conrady distortion coefficients. This dataset provides a robust foundation for training and testing deep learning models.

    \item \textit{Deep Learning-Based Camera Parameter Estimation:} We developed deep learning models based on the ResNet architecture, adapted for regression tasks to estimate camera parameters from synthetic images. The models predict intrinsic parameters using the Brown-Conrady lens distortion model, suitable for cameras with moderate distortion, in contrast to DeepCalib \cite{calib}, which utilizes the Unified Spherical Model (USM) for fisheye lenses. Innovations such as integrating image size into model training improve generalization across different image sizes and aspect ratios, while a novel distortion parameter generation method enhances the realism of synthetic data.

    \item \textit{Extensive Validation and Generalization Testing:} The models were evaluated using both the KITTI dataset \cite{kitti} and synthetic data, demonstrating strong generalization capabilities from synthetic to real-world scenarios. This work explores strategies to bridge the gap between synthetic data and practical deployment, ensuring robust model performance on real camera data.
\end{enumerate}

The rest of this paper is organized as follows. Section II provides the required background information and reviews the literature. Section III details the dataset generation steps. Section IV describes the methodology used for developing the deep learning models. Section V presents the results and discusses the model performance. Section VI concludes the paper and outlines directions for future work.

\section{Background Knowledge}

This section provides an overview of essential concepts for predicting camera calibration and distortion parameters using deep learning and synthetic data. It covers camera imaging technology, the Brown-Conrady distortion model, calibration processes, the AILiveSim simulator, the DeepCalib study, and the KITTI dataset's role in validating our methodologies.

\subsection{Camera Imaging Technology and the Brown-Conrady Distortion Model}

The pinhole camera model is a fundamental concept in camera imaging technology and represents the simplest form of a camera \cite{sturm2011camera}. It consists of a tiny hole, or aperture, through which light rays from a 3D scene pass and form an image on a plane on the opposing side of the hole. In essence, pinhole cameras have no lenses, which eliminates the complexities associated with lens-induced distortions. This simplicity makes the pinhole model particularly useful for theoretical studies in computer vision, as it adheres strictly to the principles of projective geometry. By avoiding the distortions caused by lenses, the pinhole model allows for straightforward mathematical calculations without the need for complex correction equations.

The simplicity of the pinhole camera model lies in its ability to create sharp images without the use of lenses. By allowing only a narrow beam of light through the small aperture, the model eliminates most aberrations and produces a clear image. However, this simplicity also limits light intake, resulting in darker images and requiring longer exposure times for adequate brightness.

\begin{figure}[!t]
\centering
\includegraphics[width=2.5in]{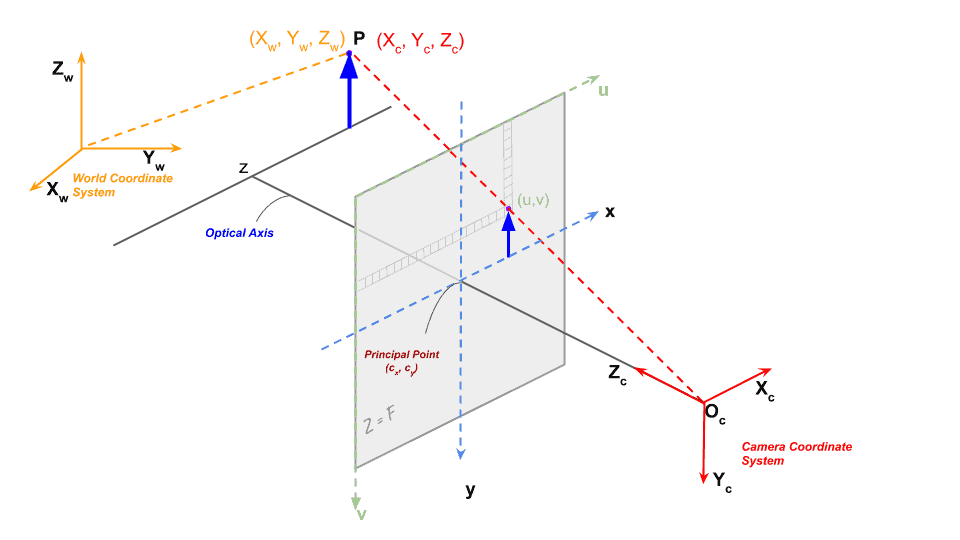}
\caption{Projection of point P onto the image plane using the Pinhole Camera Model \cite{sturm2011camera}.}
\label{fig:pinhole_camera}
\end{figure}

Figure \ref{fig:pinhole_camera} illustrates the projection of a point \(P\) onto the image plane using the pinhole camera model. The model is often referred to as a rectilinear camera model because it maintains straight lines in the 3D scene as straight lines in the 2D image.

The transformation from 3D world coordinates \((X_w, Y_w, Z_w)\) to 2D image coordinates \((u, v)\) involves several steps:

\begin{enumerate}
    \item World to Camera Coordinates:

        \begin{equation}
        \begin{bmatrix}
        X_c \\
        Y_c \\
        Z_c
        \end{bmatrix}
        = 
        \mathbf{R}
        \begin{bmatrix}
        X_w \\
        Y_w \\
        Z_w
        \end{bmatrix}
        +
        \mathbf{t},
        \end{equation}

        where \(\mathbf{R}\) is the rotation matrix, and \(\mathbf{t}\) is the translation vector that positions the camera in the world coordinate system.

    \item Camera to Image Coordinates:
        
        Using the intrinsic matrix \(\mathbf{K}\), the 3D camera coordinates are projected onto the 2D image plane:
        
        \begin{equation}
        \mathbf{K} =
        \begin{bmatrix}
        f_x & \gamma & c_x \\
        0 & f_y & c_y \\
        0 & 0 & 1
        \end{bmatrix},
        \end{equation}
        
        where \(f_x, f_y\) are the focal lengths in the x and y directions, \(\gamma\) is the skew coefficient, and \(c_x, c_y\) are the principal point coordinates.
        
        The pixel coordinates \((u, v)\) are derived from:
        
        \begin{equation}
        \begin{aligned}
        u &= f_x \frac{X_c}{Z_c} + c_x, \\
        v &= f_y \frac{Y_c}{Z_c} + c_y.
        \end{aligned}
        \end{equation}

    \item Applying Distortion:

        The Brown-Conrady model is a mathematical framework used to describe how real-world lens distortions affect the image coordinates captured by a camera \cite{Brown1971}. This model attempts to quantify the relationship between the ideal (undistorted) image coordinates and the actual (distorted) coordinates as they appear due to lens imperfections.
        
        To account for lens distortion, the Brown-Conrady model modifies the ideal image coordinates, \(x_i\) and \(y_i\), to obtain the distorted coordinates, \(x_d\) and \(y_d\), as follows:
        
        \begin{equation}
        \begin{split}
        x_d &= x_i \left(1 + k_1 r^2 + k_2 r^4 + k_3 r^6\right) \\
            &\quad + \left(2p_1 x_i y_i + p_2 (r^2 + 2x_i^2)\right), \\
        y_d &= y_i \left(1 + k_1 r^2 + k_2 r^4 + k_3 r^6\right) \\
            &\quad + \left(p_1 (r^2 + 2y_i^2) + 2p_2 x_i y_i\right),
        \end{split}\label{eq:lens_distortion}
        \end{equation}
        
        where \(x_d\) and \(y_d\) are the distorted coordinates, \(x_i\) and \(y_i\) are the ideal (undistorted) coordinates, \(k_1, k_2, k_3\) are the radial distortion coefficients, and \(p_1, p_2\) are the tangential distortion coefficients.

        The first-order radial distortion \(k_1\), often referred to as barrel distortion when positive or pincushion distortion when negative, accounts for most of the distortion in camera lenses \cite{distortions}. Figure \ref{fig:barrel_distortion} and \ref{fig:pincushion_distortion} show examples of both barrel and pincushion distortions. All other Brown-Conrady parameters are set to 0.
                
        \begin{figure*}[!t]
        \centering
        \subfloat[Barrel distortion]{\includegraphics[width=0.32\linewidth]{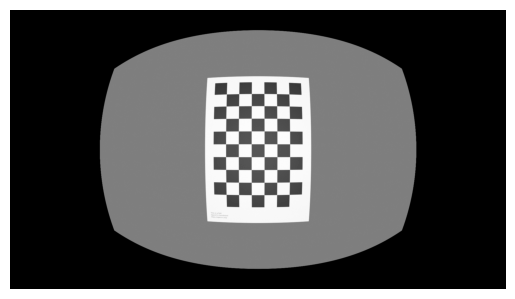}%
        \label{fig:barrel_distortion}}
        \hfil
        \subfloat[Pincushion distortion]{\includegraphics[width=0.32\linewidth]{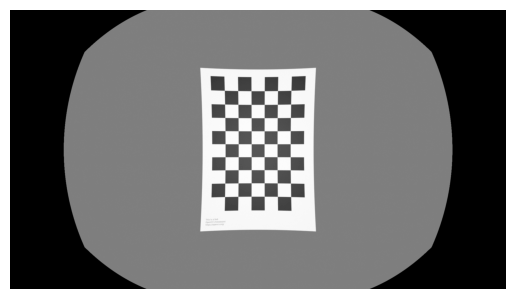}%
        \label{fig:pincushion_distortion}}
        \hfil
        \subfloat[Tangential distortion]{\includegraphics[width=0.32\linewidth]{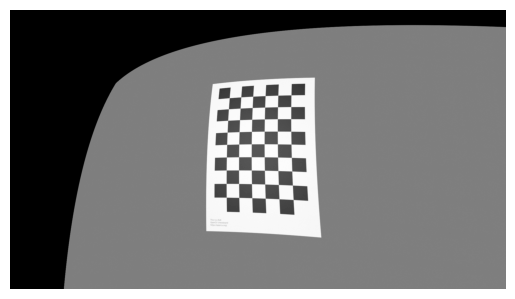}%
        \label{fig:tangential_distortion}}
        \caption{Lens distortions: (a) Barrel distortion with \(k_1 = -0.5\), (b) Pincushion distortion with \(k_1 = 0.5\), and (c) Tangential distortion with \(p_1\) and \(p_2\) set to 0.1.}
        \label{fig:combined_distortions}
        \end{figure*}

        Tangential distortion occurs when the lens and the image sensor are not perfectly
        aligned. This misalignment causes the image to appear tilted or stretched. Figure \ref{fig:tangential_distortion} illustrates the distortion of an image containing a chessboard when the tangential distortion parameters \( p_1 \) and \( p_2 \) are both set to 0.1, while all other Brown-Conrady parameters are set to zero.

\end{enumerate}

\subsection{AILiveSim Simulator}

AILiveSim is an advanced simulation platform used for generating synthetic data, which is crucial for training and validating AI systems in environments where real data collection is challenging \cite{jerome}. It allows precise manipulation of various camera settings, such as FOV, resolution, and distortion coefficients, to create realistic datasets. Adjusting aspects of the virtual environment, including background elements and scene composition, enables the simulation of diverse scenarios, enhancing the robustness of AI models trained with this data

\subsection{Evolution of Camera Calibration Techniques: Traditional Methods and Deep Learning Approaches}

Camera calibration is a fundamental process in computer vision, necessary for determining the \textit{intrinsic} and \textit{extrinsic} parameters that are vital for accurate 3D reconstruction and image rectification \cite{1087109}. Intrinsic parameters, such as focal length and principal point, describe the internal characteristics of the camera, while extrinsic parameters define its position and orientation relative to the world coordinate system.

Traditional calibration methods, such as checkerboard pattern calibration \cite{checkerboard}, Direct Linear Transformation (DLT) \cite{dlt}, and a widely used calibration technique based on projection error minimization \cite{zhang}, are well-established techniques for estimating camera parameters. These methods typically involve capturing multiple images of a known calibration pattern (e.g., a checkerboard) from different orientations. The images are then used to compute both intrinsic and extrinsic parameters by minimizing projection errors \cite{zhang}. Building on these traditional methods, Chen and Wang \cite{chen2009dynamic} proposed an efficient dynamic calibration approach for multi-camera systems, which addresses the limitations of static calibration by enabling real-time updates to calibration parameters in changing environments. Their work highlights the importance of adapting calibration techniques for applications requiring continuous monitoring and adjustment, such as surveillance and robotics.

Many current calibration techniques for cameras present challenges. These approaches often require multiple images of a calibration object, such as a checkerboard \cite{Gasparini2009, MeiRives2007, Scaramuzza2006}, dot grid \cite{ShahAggarwal1994}, or spherical object \cite{YingZha2008}. Alternatively, some methods depend on identifying specific scene features, like straight lines or vanishing points in geometrically structured scenes \cite{Antunes2017, BarretoAraujo2005, BrauerBurchardt2001, Melo2013, SwaminathanNayar2000, YingHu2004}. Another approach is based on calculating camera motion from several images \cite{Fitzgibbon2001, Kang2000, MicusikPajdla2003, XiongTurkowski1997, Zhang1996}. Among these techniques, the checkerboard-based calibration \cite{Gasparini2009, MeiRives2007, Scaramuzza2006} is the most commonly used, which involves taking multiple photos of a checkerboard. Martins et al. \cite{Martins2020} demonstrated that while checkerboard calibration is robust, it requires controlled environments, making it less suitable for dynamic scenarios like autonomous driving. Meanwhile, Barazzetti et al. \cite{Barazzetti2012} proposed a targetless approach that leverages environmental features, showing that it can perform comparably well in less controlled settings but still falls short in terms of accuracy.

Recent efforts have focused on refining traditional calibration methods to address these challenges. For instance, Huang et al. \cite{huang2021} proposed a projector-camera calibration system that leverages graph-theory-based correspondence algorithms and bundle adjustment to jointly optimize projector-camera parameters, addressing inaccuracies caused by imperfect calibration targets. This highlights ongoing advancements in calibration processes, even within traditional methodologies, to improve accuracy and robustness in real-world conditions.

In contrast, recent advancements in deep learning have introduced new methods for camera calibration, offering more flexibility and efficiency. DeepCalib, for instance, employs convolutional neural networks (CNNs) to estimate intrinsic camera parameters from a single image \cite{calib}, without requiring specific calibration targets or multiple images. It utilizes the Unified Spherical Model (USM) for distortion representation, making it particularly suitable for wide FOV cameras. However, its reliance on the USM limits its generalizability to cameras modeled by the Brown-Conrady distortion model, which is more applicable to linear cameras used in automotive applications for detecting distant objects with minimal distortion \cite{Bogdan2018}.

To ensure robustness across varied imaging conditions, our research emphasizes the use of the Brown-Conrady model, which, while primarily associated with linear cameras, is broadly applicable across different types of cameras within this context. We train our models on one dataset and test them on another to guarantee that they perform well across diverse imaging scenarios, thus addressing the limitations seen in models like DeepCalib.

This combined perspective highlights the evolution from traditional to modern deep learning-based calibration methods, demonstrating how each contributes to the development of more accurate and flexible camera calibration techniques.

\subsection{KITTI Dataset}

The KITTI dataset is a benchmark for autonomous driving research, providing real-world data captured from multiple sensors, including cameras and LiDAR \cite{kitti}. It is instrumental for evaluating computer vision algorithms in diverse environments and includes detailed calibration information \cite{kitti}. This makes KITTI ideal for validating models trained on synthetic data and assessing their performance in predicting camera parameters such as H-FOV, principal point, and Brown-Conrady distortion coefficients.

\section{Dataset Generation}
The dataset generation process is central to this study's approach to predicting camera calibration and distortion parameters. The Camera Parameter Search (CPS) dataset is generated using the AILiveSim simulation platform, specifically designed to replicate real-world camera behavior within a controlled virtual environment. This dataset, consisting of 1,495,000 images, is carefully crafted to facilitate both the training and testing of deep learning-based approaches for analyzing intrinsic camera parameters under various conditions. By providing a comprehensive range of scenarios, the CPS dataset ensures robust model performance and reliable calibration predictions across diverse camera setups.

\subsection{Synthetic Data Creation}
The CPS dataset is developed using AILiveSim, a robust simulation platform capable of generating high-quality synthetic images with diverse camera settings and environmental conditions. The dataset comprises 13 unique image sets, each containing 10,000 images, and spans a range of FOV settings from 30° to 150° in 10-degree increments. This range is chosen to simulate real-world scenarios where cameras operate under different angles and perspectives.

Each set of images is captured in a virtual city environment, selected due to the abundance of straight lines in urban landscapes, such as buildings and roads. These straight lines are ideal for analyzing optical distortion effects. The camera is mounted on a virtual vehicle, emulating the data collection methodology used in real-world datasets like KITTI, where cameras are mounted on cars navigating through city streets.

\subsection{Applying Distortions}
\begin{figure*}[!t]
\centering
\subfloat[]{\includegraphics[width=2.5in]{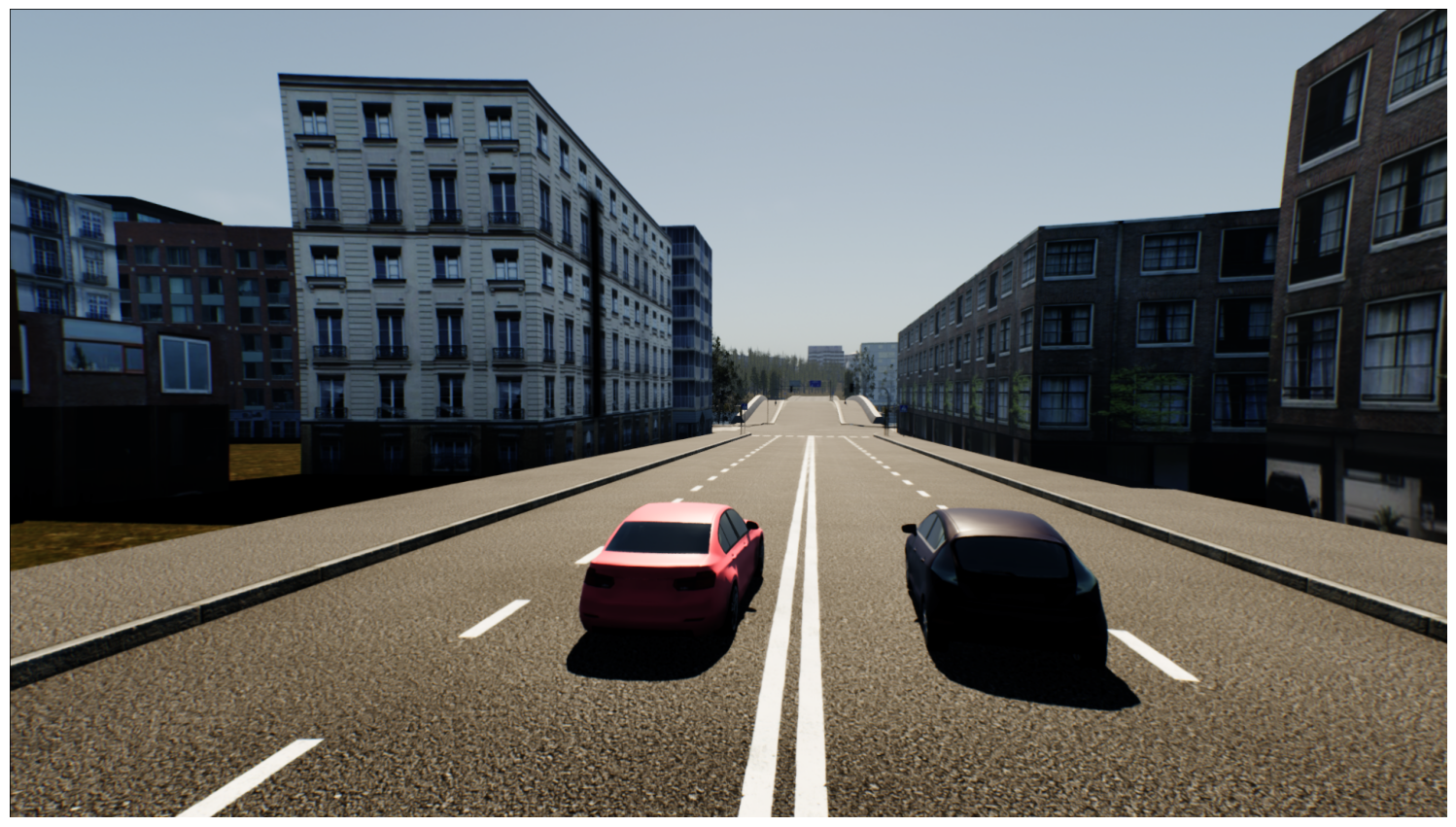}%
\label{fig:90_ALS_distortion}}
\hfil
\subfloat[]{\includegraphics[width=2.5in]{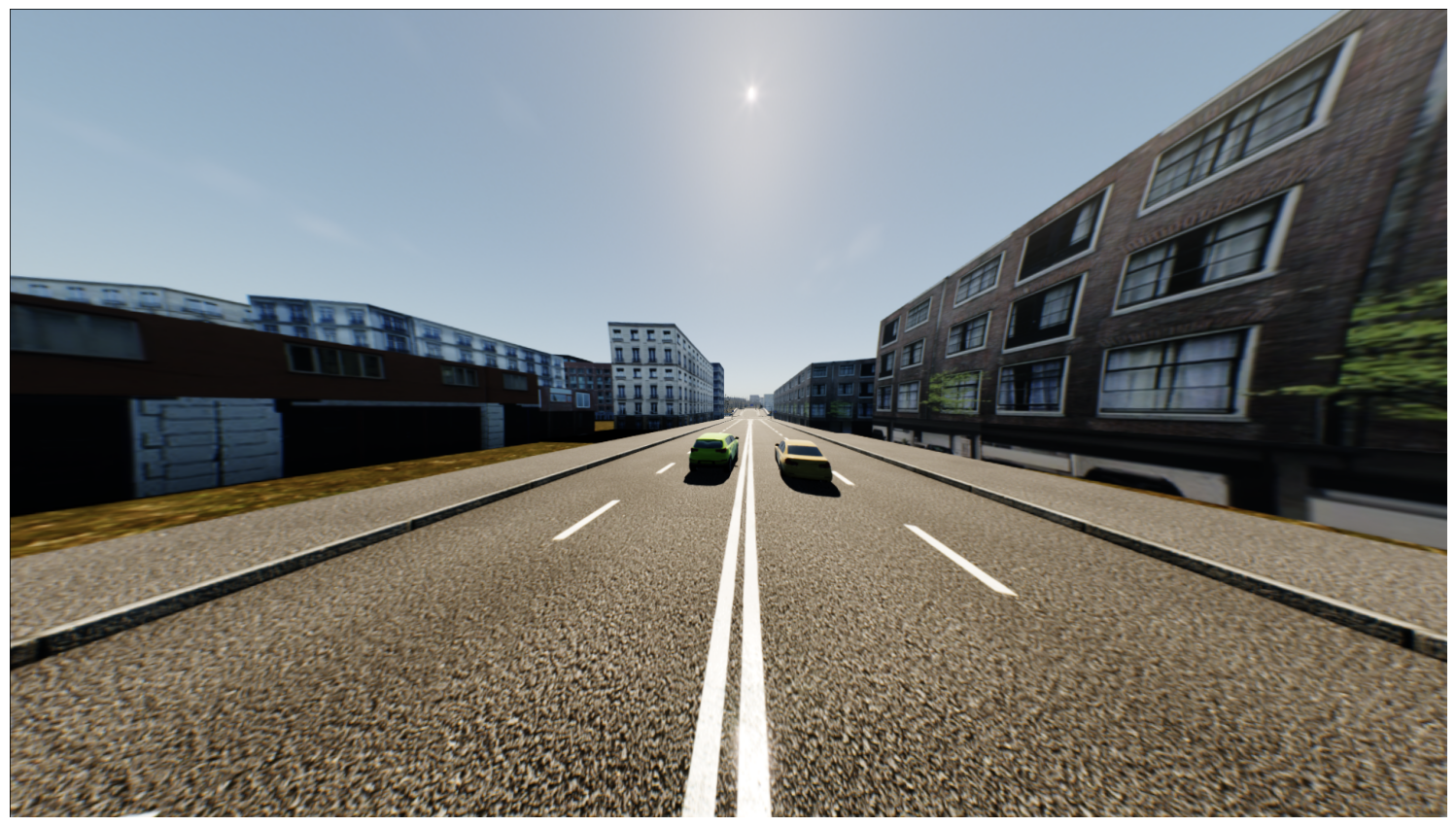}%
\label{fig:150_ALS_distortion}}
\caption{Distortion effects on a city scene from AILiveSim with different H-FOV settings. (a) 90° H-FOV with \(k_1 = 0.25\). (b) 150° H-FOV with \(k_1 = 0.25\).}
\label{fig:90_150_ALS_distortion}
\end{figure*}

The dataset's generation involves capturing images with and without distortions to establish a comprehensive training dataset for deep learning models. Initially, images are collected with no distortion to serve as a baseline. Following this, each set is processed with five distinct distortion parameter sets based on the Brown-Conrady model, incorporating radial distortion coefficients (\(k_1, k_2, k_3\)) and tangential distortion coefficients (\(p_1, p_2\)). Furthermore, the principal axes (\(c_x, c_y\)) are deliberately shifted in the x and y directions across different sets to simulate variations in camera alignment. An example of the distortion effects is illustrated in Figure \ref{fig:90_150_ALS_distortion}, which compares images with different H-FOV subjected to the same radial distortion. Figure \ref{fig:90_ALS_distortion} shows an image with a 90° H-FOV, while Figure \ref{fig:150_ALS_distortion} depicts an image with a 150° H-FOV. Both images are distorted using a first-order radial distortion coefficient of \(k_1 = 0.25\). The comparison highlights how the extent of distortion varies with changes in H-FOV, illustrating the relationship between field of view and distortion intensity.

\subsection{Adaptive Distortion Ranges}

To generate realistic camera distortions for different H-FOVs, a method is developed to determine optimal distortion parameters using the Brown-Conrady model. The goal is to simulate real-world lens distortions while striking a balance between achieving realistic effects and maintaining control over the parameters. The process involves several key steps to ensure the parameters are not arbitrarily selected but rather calculated to remain within practical limits. This is particularly important to avoid excessive distortions that would result in unrealistic images. The steps are as follows.

\begin{enumerate} 
    \item \textit{Reference Point Selection}: We choose the top-left pixel in the image as the point-of-interest (POI) that defines how much this point would move for a given set of lens distortion parameters. Since the optical axis is typically at the center of the image, lens distortions are strongest near the edges. Since we are trying to define reasonable values for the lens parameters, and not exact values, we could have chosen any other corner from the image.
    \item \textit{Maximum Distortion in Pixels}: We determine the maximum number of pixels by which the lens distortion parameters can translate the point-of-interest (POI). This sets a practical upper limit on the distortion.
    \item \textit{Choosing Values for the Lens Parameters}: We randomly choose one of the lens parameters \(k_1\), \(k_2\), \(k_3\), \(p_1\) or \(p_2\) as a starting point and set the others to zero. Using the Bisection algorithm \cite{10.1145/2701.2705}, we obtain the interval \([a, b]\) for the lens parameter where the movement induced by the distortion is within the maximum distortion in pixels mentioned in the previous step. 
    We use Equation \ref{eq:lens_distortion} in the Bisection algorithm for calculating the movement caused by the distortion. Without loss of generality, the final value for the lens parameter is chosen from a Uniform distribution \(\mathcal{U}(a, b)\). We repeat this procedure until we have defined values for all the lens parameters.
\end{enumerate}

Our findings show that different H-FOV settings require carefully chosen distortion parameters due to their unique sensitivities to distortion. The study also considers additional distortion parameters from the Brown-Conrady model, including \(k_2\), \(k_3\), \(p_1\), \(p_2\), and adjustments to the principal axes (\(c_x\), \(c_y\)). These are randomly selected and optimized using root-finding techniques to achieve realistic image modifications. 

To manage distortions effectively and avoid model biases, we modify the focal length of the camera in order to exclude black pixels caused by distortions outside of the camera's FOV.

\section{Methodology}
\subsection{Model Architecture}
The Deep-BrownConrady (DBC) model utilizes a ResNet50 architecture \cite{7780459}, modified for regression tasks to predict eight camera parameters: three radial distortion coefficients (\(k_1, k_2, k_3\)), two tangential distortion coefficients (\(p_1, p_2\)), two principal axes shifts (\(c_x, c_y\)), and the FOV. The model was adapted by replacing the final classification layers with fully connected layers suited for regression, followed by a regression head layer for continuous output.

\subsection{Training Procedure}
The training process utilized a Mean Squared Error (MSE) loss function to minimize the difference between predicted and actual parameter values. The AdamW optimizer, known for its effective handling of weight decay \cite{loshchilov2019decoupled}, was employed with a batch size of 128 and an initial learning rate of 0.01. The learning rate was gradually decayed over two milestones at 20 and 40 epochs to ensure stable convergence. Each model was trained for 60 epochs, with the dataset split into 70\% training, 15\% validation, and 15\% testing, ensuring comprehensive model evaluation and adjustment.

Starting with Deep-BrownConrady v1 (DBC v1), which served as the baseline model, the architecture was based on ResNet50, adapted for regression tasks to predict camera parameters. This baseline model laid the foundation for further improvements.

In Deep-BrownConrady v2 (DBC v2) and later versions, the training process was enhanced by incorporating images of two different resolutions (1920x1080 and 1392x512 pixels). This change aimed to improve the model's ability to generalize across different image sizes and aspect ratios, reflecting real-world scenarios where camera images often vary in resolution and aspect ratio. The introduction of a batch-sampler was crucial in this context, allowing for batches of uniformly sized images during training. This approach helped mitigate any biases that could result from processing images of mixed resolutions together, ensuring the model focused on learning robust, resolution-independent features. As a result, the model's ability to infer geometric properties improved, especially in diverse image contexts, enhancing its predictive accuracy for camera parameters.

Building on these improvements, Deep-BrownConrady v3 (DBC v3) further refined the architecture by incorporating the original image size into the final feature vector, with the values normalized to avoid issues arising from different numerical magnitudes. By feeding the normalized image size into the fully connected layer, the model gains additional geometric information about the camera, which can be utilized during the prediction of camera intrinsic parameters. This helps the model better understand the relationship between image scale and the camera's geometric properties, leading to improved accuracy in tasks such as camera parameter estimation, where the distinction between intrinsic scene properties and image scale is crucial.

\begin{figure*}[!t]
\centering
\includegraphics[width=\textwidth]{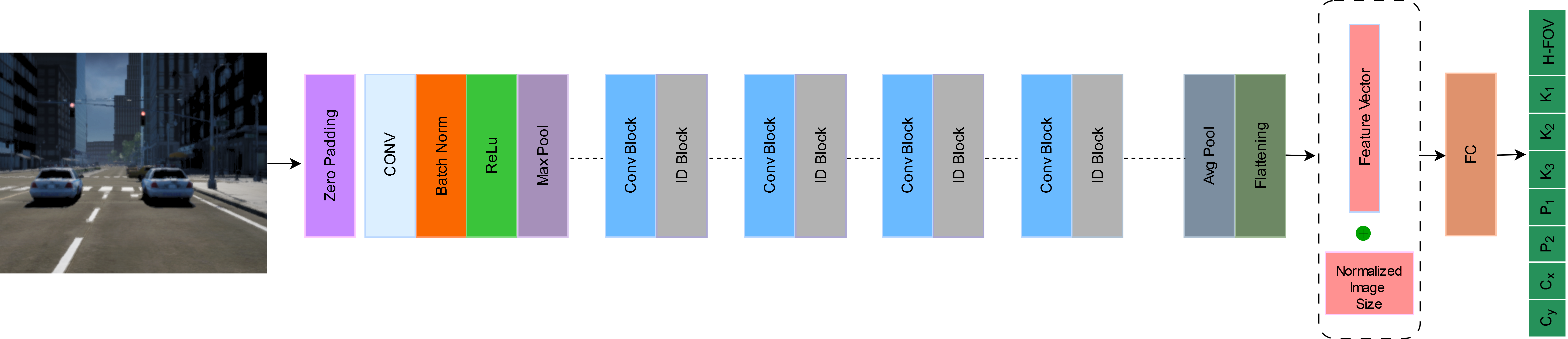}
\caption{Architecture of the DBC v3 model, based on the ResNet50 framework.}
\label{fig:resnet_architecture}
\end{figure*}

\section{Results and Discussion}

This section presents a comprehensive evaluation of the camera parameter extraction models, developed iteratively from DBC v1 to DBC v3. Each model was systematically trained on synthetic datasets and tested on both synthetic transformations and real-world data from the KITTI dataset. The aim was to assess and improve the models' robustness and accuracy in predicting camera calibration parameters.

\subsection{Model Development and Performance Analysis}
\label{subsec:model_dev_performance}

The models were developed using the ResNet50 architecture, trained on varying datasets as detailed in section III, and fine-tuned with specific adaptations, as discussed in section IV, to enhance their predictive capabilities. The primary outputs were camera parameters such as H-FOV, principal points, and Brown-Conrady distortion coefficients (\(k_1, k_2, k_3, p_1, p_2\)).

To evaluate the robustness of each model, various image transformations were applied to the test set, including Gaussian blur, motion blur, random gamma correction and random pixel dropout. This is important because real-world applications often involve images that are not perfectly captured, and the ability of a model to maintain accuracy under such conditions is critical for reliable camera parameter estimation. Table \ref{table_model_performance} summarizes the performance of each model across different transformations, with the 'wo' column referring to performance without any transformation, meaning the original, unmodified test images.

\begin{table*}[!t]
\centering
\caption{MSE loss values of DBC models across different transformations (wo: without transformation).}
\begin{tabular}{|c|c|c|c|c|c|}
\hline
\textbf{Model} & \textbf{wo} & \textbf{Gaussian Blur} & \textbf{Motion Blur} & \textbf{Random Gamma} & \textbf{Dropout} \\
\hline
DBC v1 & 31.4e-03 & 31.9e-03 & 31.7e-03 & 31.4e-03 & 31.6e-03 \\
DBC v2 & 13.6e-03 & 13.8e-03 & 13.9e-03 & 13.7e-03 & 13.9e-03 \\
DBC v3 & 3.36e-03 & 3.38e-03 & 3.39e-03 & 3.37e-03 & 3.39e-03 \\
\hline
\end{tabular}
\label{table_model_performance}
\end{table*}

As shown in Table \ref{table_model_performance}, the MSE loss values significantly decreased across the model versions, with DBC v3 demonstrating the lowest loss values across all transformations. This steady reduction in MSE highlights the increasing robustness of the later models in handling complex image transformations, such as Gaussian blur, motion blur, and random gamma adjustments.

The substantial improvement in loss values from DBC v1 to DBC v3 reflects the enhanced capability of the newer models to generalize better to various distortions and image conditions. The introduction of fine-tuned adaptations in later versions allowed the models to learn more robust, transformation-independent features, resulting in more accurate predictions of camera parameters even in challenging scenarios. Thus, the decreasing MSE values serve as an indication of the progressive improvements in model performance and reliability.

To evaluate the spatial accuracy of the predicted camera parameters, an error map was generated by simulating the distortion and undistortion process. A blank image of size \(1392 \times 512\) was used, and a grid of pixel coordinates was created to represent the entire image. These coordinates were first distorted using the true parameters from the test set, simulating the real-world effects of lens distortion. The distorted coordinates were then undistorted using the predicted parameters from each model (DBC v1, DBC v2, and DBC v3). For every pixel, the error was calculated as the Euclidean distance between its original position and its final position after distortion and undistortion. These errors were normalized by the image width and visualized as an error map, offering a detailed representation of spatial deviations.

\begin{figure*}[!t]
\centering
\subfloat[Top Horizontal Line Errors]{%
    \includegraphics[width=0.32\textwidth]{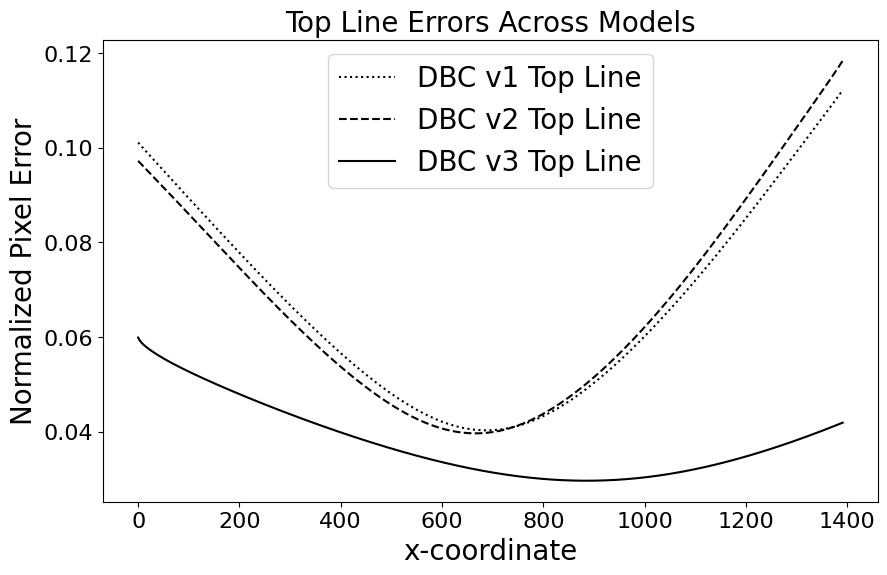}%
    \label{fig:top_line_errors}
}
\hfill
\subfloat[Center Horizontal Line Errors]{%
    \includegraphics[width=0.32\textwidth]{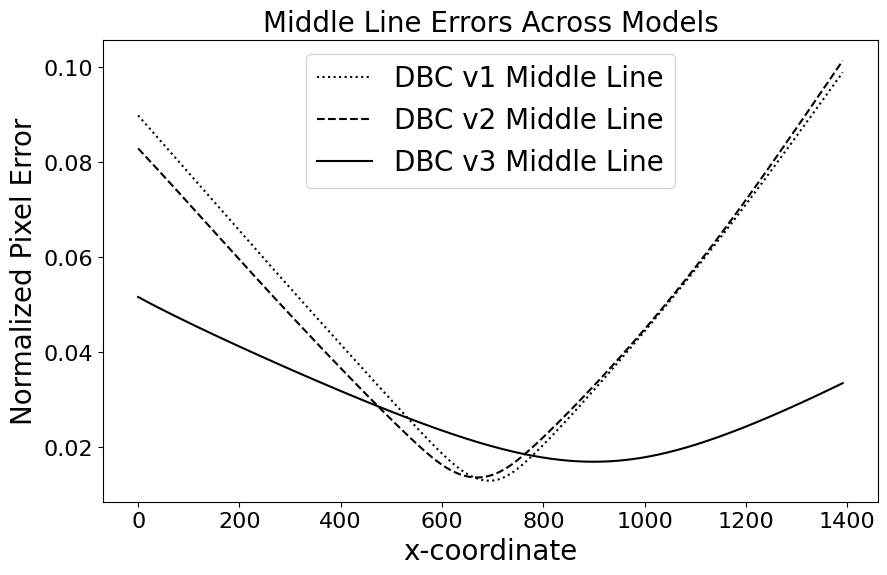}%
    \label{fig:middle_line_errors}
}
\hfill
\subfloat[Bottom Horizontal Line Errors]{%
    \includegraphics[width=0.32\textwidth]{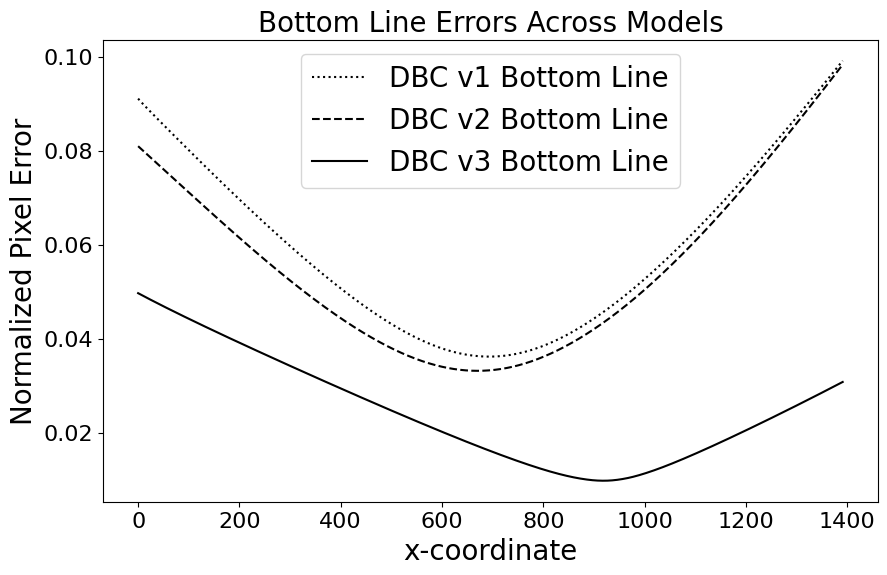}%
    \label{fig:bottom_line_errors}
}
\caption{Normalized pixel-wise errors along top, middle, and bottom horizontal lines for DBC v1, DBC v2, and DBC v3. Each graph shows normalized pixel-wise errors across x-coordinates for all three models.}
\label{fig:line_errors_comparison}
\end{figure*}

This process was repeated for a subset of 5000 randomly selected test images, and a mean error map was generated to summarize the overall spatial performance of the models. From this mean error map, three horizontal lines—top, middle, and bottom—were extracted for further analysis. These lines provide insight into how the models handled distortions in different regions of the image. The normalized pixel errors along these lines are visualized in Figure \ref{fig:line_errors_comparison}, where the performance of the three models is compared.

\subsection{Comparison with Traditional Calibration Methods}

To evaluate the performance of our models (DBC v1, DBC v2, and DBC v3) compared to traditional camera calibration methods, we conducted an experiment using different approaches. For the traditional method, varying numbers of images of a known calibration pattern were provided as input. In contrast, the Deep-BrownConrady methods used a single image captured in the wild, without any visible objects of known sizes or shapes. The same distortion parameters were applied consistently across all images.

Table \ref{tab:error_table} presents the comparison of the normalized pixel-wise errors for the traditional calibration approach and the DBC models. Each method’s performance is evaluated along the top, middle, and bottom horizontal lines of the error map. The error map was generated using the distortion and undistortion process described in Section \ref{subsec:model_dev_performance}. Briefly, it involves simulating distortions using true parameters, then undistorting with predicted parameters, and calculating the normalized Euclidean error for every pixel. This error map provides a detailed spatial representation of deviations across the image.

The table demonstrates that the traditional calibration method struggles to match the accuracy of any of our DBC models when using only a few images. With a low number of images, such as 1, 2, or 3, the accuracy of the results depends significantly on the location and orientation of the calibration pattern in the image. Specifically, when the calibration object was primarily fronto-parallel to the camera, the results were more accurate. These findings align with expectations: as more information becomes available for camera calibration, reducing uncertainty, better calibration results are achieved.

This trend highlights the trade-off between the dependency on structured calibration objects in the traditional method and the ability of our DBC models to perform calibration directly from distorted images without such objects or multiple images. While the traditional approach benefits from an increased number of calibration images, our DBC models maintain consistent performance across all scenarios, demonstrating their applicability in dynamic environments where calibration objects or multiple views are unavailable.

This comparison underscores the limitations of traditional calibration techniques when restricted to a limited number of calibration images and highlights the efficacy of our deep learning-based approach, which eliminates the need for pre-defined objects or multiple images for calibration.

\begin{table*}[t]
    \centering
    \caption{Comparison of the traditional camera calibration approach with DBC v1, DBC v2, and DBC v3: The traditional calibration method relied on images of a known calibration object, whereas the Deep-BrownConrady models utilized a single image captured in the wild, without any objects of known shape or size.}
    \label{tab:error_table}
    \begin{tabular}{|l|c|c|c|c|c|c|}
        \hline
        \textbf{Method} & \multicolumn{2}{|c|}{\textbf{TOP HORIZONTAL LINE}} & \multicolumn{2}{|c|}{\textbf{MIDDLE HORIZONTAL LINE}} & \multicolumn{2}{|c|}{\textbf{BOTTOM HORIZONTAL LINE}} \\
        \hline
        & \textbf{MIN ERROR} & \textbf{MAX ERROR} & \textbf{MIN ERROR} & \textbf{MAX ERROR} & \textbf{MIN ERROR} & \textbf{MAX ERROR} \\
        \hline
        DBC v1 & 34.96e-03 & 129.66e-03 & 9.01e-03 & 108.23e-03 & 9.01e-03 & 116.89e-03 \\
        \hline
        DBC v2 & 13.47e-03 & 70.92e-03 & 16.68e-03 & 56.18e-03 & 16.68e-03 & 62.98e-03 \\
        \hline
        DBC v3 & 5.84e-03 & 41.20e-03 & 23.65e-03 & 44.12e-03 & 23.65e-03 & 49.73e-03 \\
        \hline
        Calibration, 1 image & 186.13e-03 & 753.72e-03 & 122.22e-03 & 767.03e-03 & 203.73e-03 & 776.89e-03 \\
        \hline
        Calibration, 3 images & 0.09e-03 & 33.22e-03 & 0.39e-03 & 26.08e-03 & 0.18e-03 & 37.48e-03 \\
        \hline
        Calibration, 5 images & 2.97e-03 & 27.36e-03 & 3.31e-03 & 22.93e-03 & 2.93e-03 & 30.25e-03 \\
        \hline
        Calibration, 7 images & 5.60e-03 & 12.42e-03 & 5.42e-03 & 10.96e-03 & 5.57e-03 & 13.77e-03 \\
        \hline
        Calibration, 9 images & 2.69e-03 & 9.31e-03 & 3.69e-03 & 7.70e-03 & 2.05e-03 & 10.74e-03 \\
        \hline
    \end{tabular}
\end{table*}

\subsection{Key Improvements and Insights}

The iterative development of the models led to two key improvements:

\begin{itemize}
    \item \textit{Handling Aspect Ratio Variability:} By introducing image dimensions into the model's feature vector (as in DBC v3), we addressed the challenge of varying image sizes and shapes, leading to substantial performance gains.
    \item \textit{Real-World Applicability:} The progressive reduction in loss values on the KITTI dataset confirmed the models' growing ability to generalize beyond synthetic training data, validating their applicability in diverse real-world conditions.
\end{itemize}

To further illustrate the models' ability to predict and correct distortions, a synthetic test image containing horizontal and vertical lines was generated. This image was distorted using the true Brown-Conrady parameters from the test set. The distorted image was then undistorted using the predicted parameters from all three models: DBC v1, DBC v2, and DBC v3.

\begin{figure}[!th]
\centering
\subfloat[Original]{\includegraphics[width=2in]{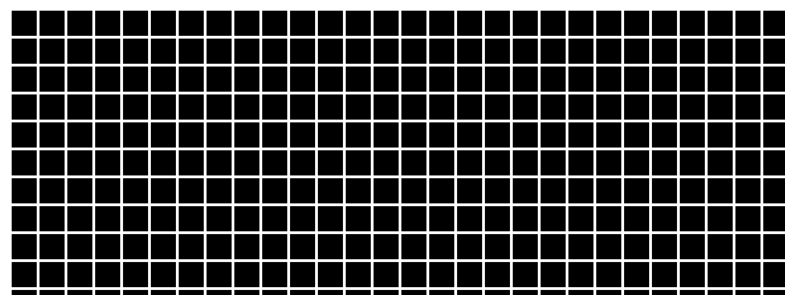}}\\[1em]
\subfloat[Distorted]{\includegraphics[width=2in]{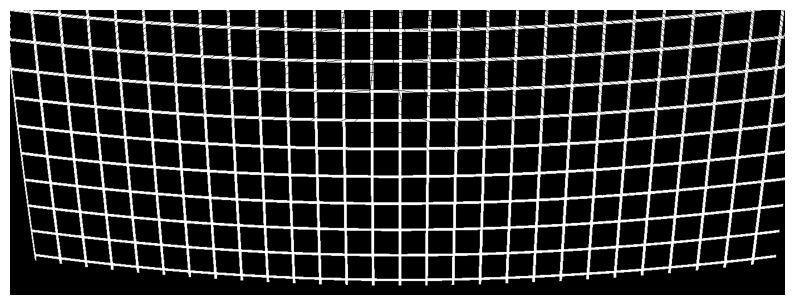}}\\[1em]
\subfloat[DBC v1]{\includegraphics[width=2in]{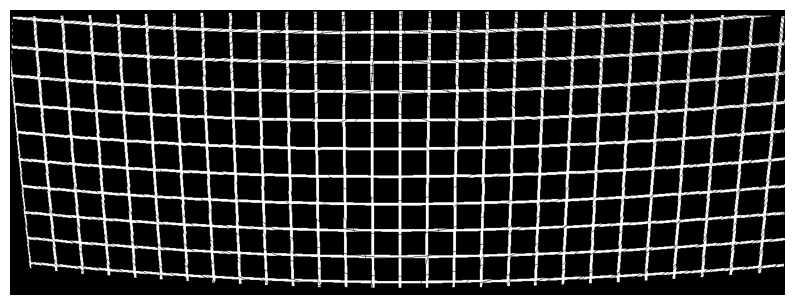}}\\[1em]
\subfloat[DBC v2]{\includegraphics[width=2in]{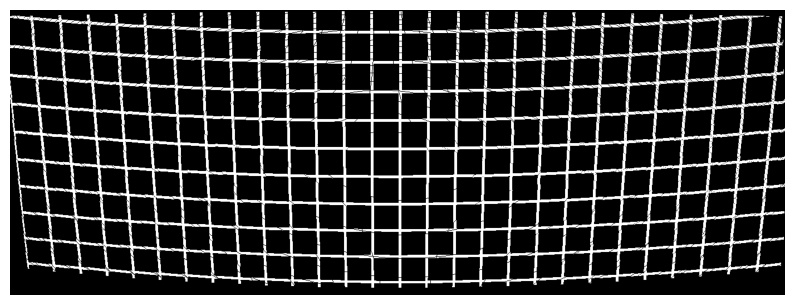}}\\[1em]
\subfloat[DBC v3]{\includegraphics[width=2in]{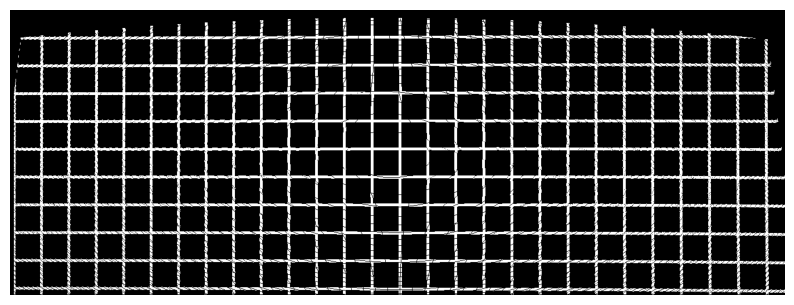}}
\caption{Visualization of the distortion and undistortion process. (a) Original line image, (b) Image distorted using true distortion parameters, (c) Undistorted image using predicted parameters from DBC v1, (d) Undistorted image using DBC v2, (e) Undistorted image using DBC v3.}
\label{fig_distortion_vertical}
\end{figure}

The visual results of this experiment are shown in Figure \ref{fig_distortion_vertical}. These images demonstrate how the predicted parameters from each model progressively improved in their ability to reverse the distortion, with DBC v3 producing the most accurate undistorted image.

The use of straight lines as a test case was critical for visualizing and understanding the distortion effects, as well as evaluating the correction accuracy of each model. The results show that the predicted camera parameters from each model effectively reverse the distortions, with DBC v3 achieving the most precise restoration of the original straight-line image.

\FloatBarrier
\section{Conclusions and Future Work}

This research demonstrated the feasibility and effectiveness of using deep learning models for predicting camera calibration and distortion parameters from a single image, leveraging synthetic data. Through iterative model development DBC v1  to DBC v3, we progressively refined our approach, culminating in a model that achieved near-perfect predictions with minimal loss. The key enhancement involved modifying the ResNet architecture to incorporate image size information directly into the feature vector, significantly improving the model's ability to predict calibration parameters accurately by linking image dimensions with parametric outputs.

Our findings confirmed that deep learning models could reliably estimate camera parameters even in the presence of various distortions and varying FOVs. We identified a strong relationship between FOV and image distortion, noting that larger FOVs with even minor distortions significantly affect image quality. Furthermore, the modifications to the ResNet architecture, including the use of normalized image dimensions, proved effective in addressing the challenges posed by varying image geometries.

Despite these advancements, the study faced several limitations, such as the scarcity of publicly available datasets with comprehensive calibration data and the substantial computational resources required for training. These constraints limited the scope for extensive validation and experimentation. However, our approach of generating synthetic data with characteristics similar to real-world datasets like KITTI demonstrated promising generalization capabilities, highlighting the potential of synthetic data in real-world applications.

Future work could explore incorporating additional image information, such as edge detection and pixel clustering, to improve the model's geometric feature learning. Additionally, using graph-based models like Graph Neural Networks (GNNs) may enhance predictions by capturing spatial relationships between pixels. Expanding the training dataset to include a wider range of aspect ratios would improve model generalization, and exploring efficient training techniques like model compression or knowledge distillation could reduce computational costs while maintaining performance.

\bibliographystyle{IEEEtran}

\newpage

\vfill
\end{document}